\documentclass[runningheads]{llncs}

\usepackage{multirow} 
\usepackage{makecell}
\usepackage{caption}
\usepackage{floatrow}
\newfloatcommand{capbtabbox}{table}[][\FBwidth]
\usepackage{eccv}

\usepackage{eccvabbrv}

\usepackage{graphicx}
\usepackage{booktabs}

\usepackage[accsupp]{axessibility}  % Improves PDF readability for those with disabilities.

\usepackage{hyperref}

\usepackage{orcidlink}

\begin{document}

\title{CEIA: CLIP-Based Event-Image Alignment for Open-World Event-Based Understanding}

\titlerunning{CEIA: CLIP-Based Event-Image Alignment}

\author{Wenhao Xu\and
Wenming Weng\and
Yueyi Zhang\and
Zhiwei Xiong}



% Springer Heidelberg, Tiergartenstr.~17, 69121 Heidelberg, Germany
\institute{
University of Science and Technology of China}

\maketitle

\begin{abstract}
% One way to achieve event-text alignment is to train a large event-text model. However, it poses a huge challenge due to the shortage of paired event-text data. 

We present CEIA, an effective framework for open-world event-based understanding. Currently training a large event-text model still poses a huge challenge due to the shortage of paired event-text data. In response to this challenge, CEIA learns to align event and image data as an alternative instead of directly aligning event and text data. Specifically, we leverage the rich event-image datasets to learn an event embedding space aligned with the image space of CLIP through contrastive learning. In this way, event and text data are naturally aligned via using image data as a bridge. Particularly, CEIA offers two distinct advantages. First, it allows us to take full advantage of the existing event-image datasets to make up the shortage of large-scale event-text datasets. Second, leveraging more training data, it also exhibits the flexibility to boost performance, ensuring scalable capability. In highlighting the versatility of our framework, we make extensive evaluations through a diverse range of event-based multi-modal applications, such as object recognition, event-image retrieval, event-text retrieval, and domain adaptation. The outcomes demonstrate CEIA's distinct zero-shot superiority over existing methods on these applications.

\end{abstract}

\begin{figure}[t!]
\centering
  \begin{subfigure}{0.5\linewidth}
    \centering
    \includegraphics[width=6cm]{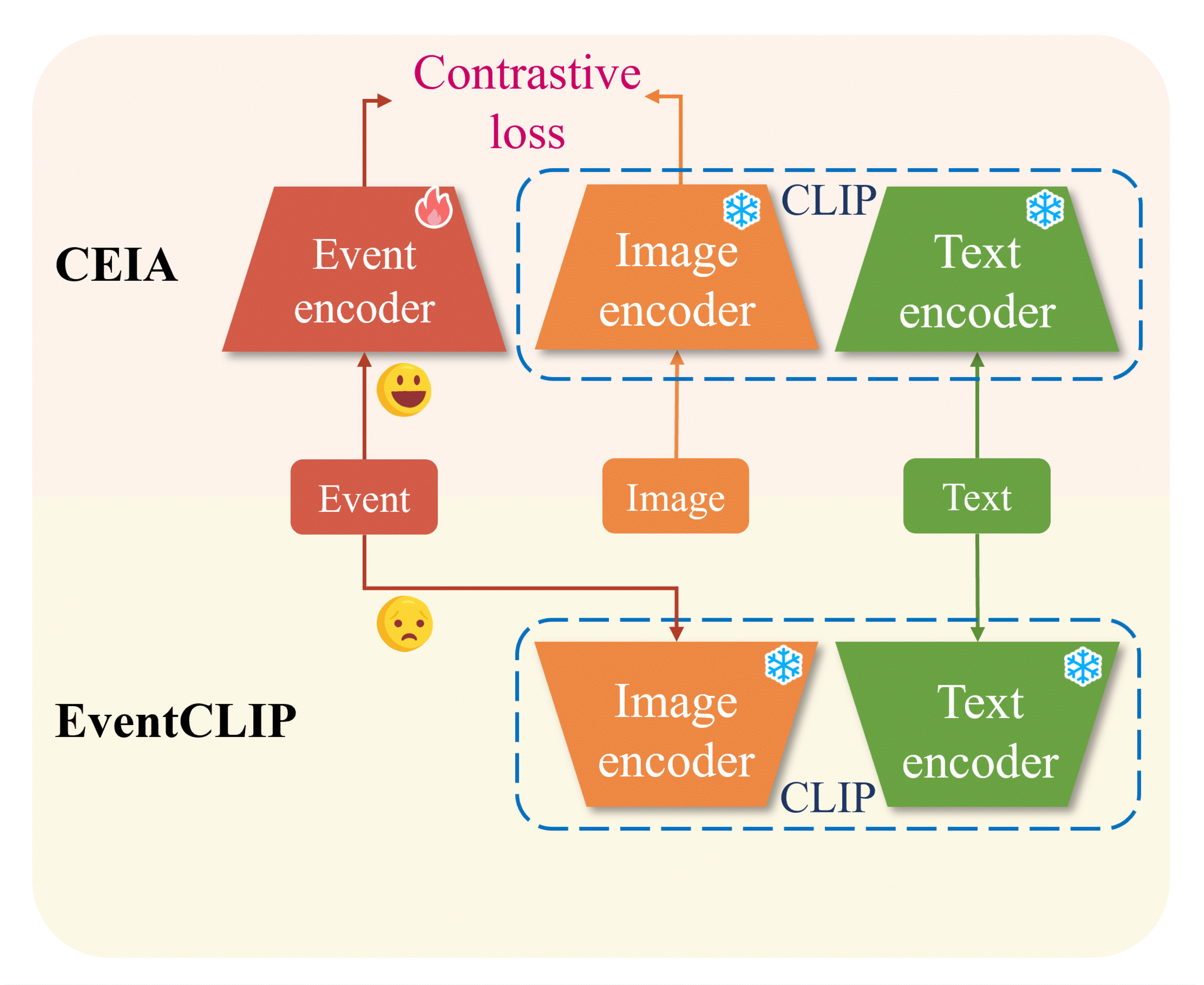}
    % \vspace{-2mm}
    \caption{EventCLIP \cite{wu2023eventclip} \textit{VS.} Our CEIA.}
    \label{fig:EventCLIP}
  \end{subfigure}
  \begin{subfigure}{0.49\linewidth}
    \centering
    \includegraphics[width=5.98cm]{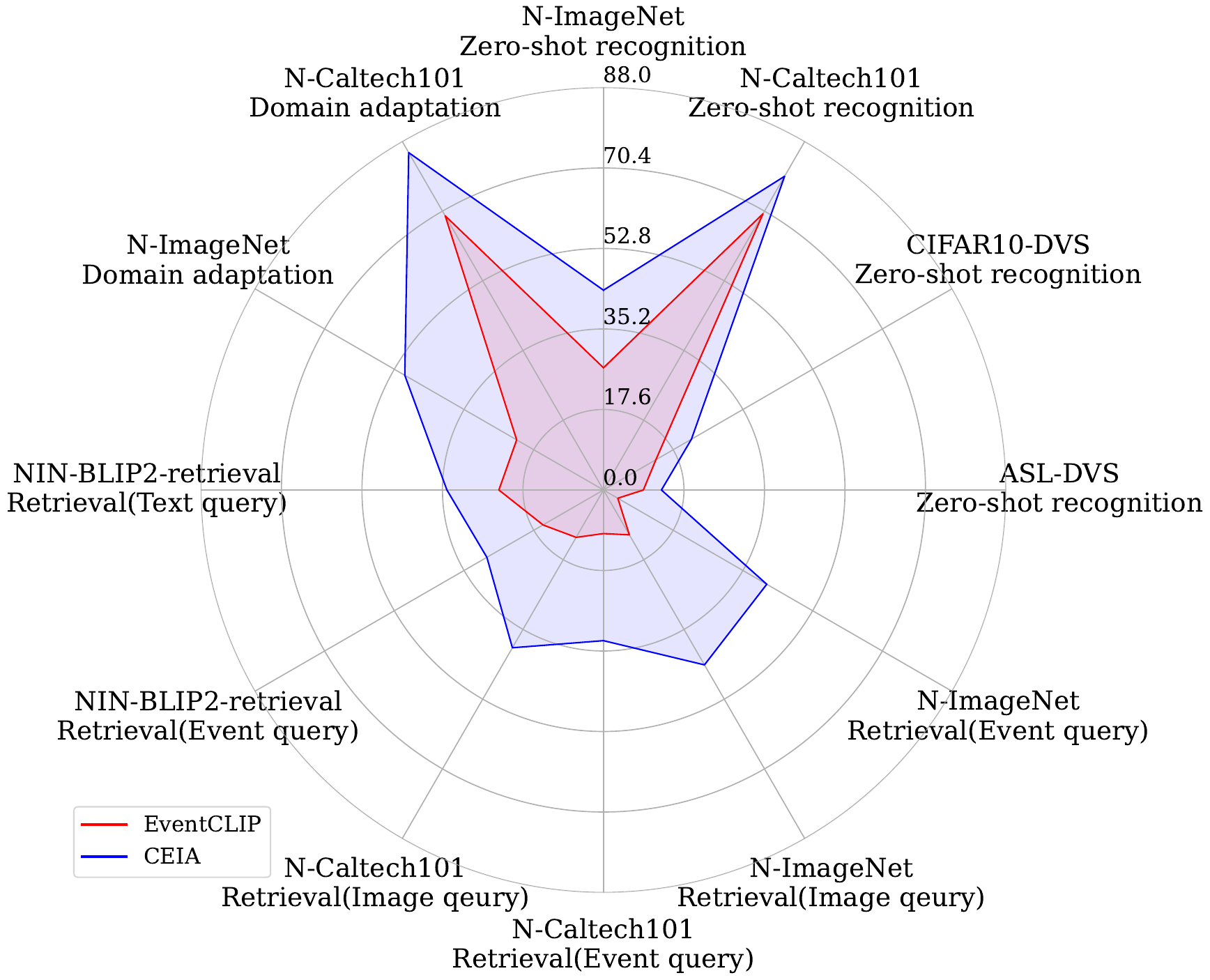}
    % \vspace{-2mm}
    \caption{Comparison of our CEIA and EventCLIP \cite{wu2023eventclip} on various datasets and tasks.}
    \label{fig:ELIP}
  \end{subfigure}
  % \vspace{-8mm}
  \caption{(a) Compared with EventCLIP \cite{wu2023eventclip} that directly utilizes the frozen CLIP's image encoder, our CEIA learns an event encoder to alleviate the event-image modality disparity. (b) Comparison of our CEIA and EventCLIP \cite{wu2023eventclip} on various datasets and tasks. For zero-shot recognition and domain adaptation, we report Acc1 (\%), while for event-image retrieval and event-text retrieval, we report R@1 (\%) \cite{lee2018stacked}.
  }
  \label{fig:ELIPELIP}
  % \vspace{-2mm}
  
\end{figure}

\section{Introduction}
\label{sec:intro}
Event cameras are sensors that asynchronously measure the intensity changes at each pixel independently with microsecond temporal resolution \cite{gallego2020event}. Compared to conventional frame cameras, event cameras exhibit several exceptional advantages. They have a very high dynamic range, are immune to motion blur, and provide measurements with a microsecond-level temporal resolution. These inherent advantages have sparked considerable interest in event cameras, notably for computer vision applications such as autonomous navigation \cite{kim2022beyond}, robotics \cite{glover2018controlled}, and virtual reality (VR) \cite{li2023track}.

Despite the superiority of event cameras, event-based algorithms are still in their infancy, facing two major issues: the shortage of large-scale datasets and the failure of modeling new data distributions in the real world. Consequently, it is imperative to explore the zero-shot event-based algorithms. Very recently, several works \cite{wu2023eventclip,zhou2023clip} have explored how to transfer impressive zero-shot knowledge from CLIP \cite{radford2021learning} to event-based vision. 
EventCLIP \cite{wu2023eventclip} demonstrated the feasibility of improving event-based zero-shot capability by first transforming events into frames and then directly utilizing frozen CLIP to extract event features. However, the image encoder of CLIP is primarily trained on natural images, resulting in a significant domain gap between images and the transformed frames. Therefore, the performance is severely impeded. To address this shortcoming, in this paper, we propose CEIA, an effective framework to adapt CLIP to event data while accommodating a wide range of open-world event-based understanding tasks.

CEIA achieves its goal by learning an individual event encoder through cross-modal contrastive learning, instead of directly utilizing the frozen image encoder like EventCLIP. The differences between EventCLIP and CEIA are depicted in \cref{fig:EventCLIP}. In particular, we observe that, unlike 3D point clouds \cite{guo2020deep} or depth maps, event data have a notable characteristic: they are often accompanied by available paired image data. This accessibility is largely thanks to the widespread use of dynamic and active-pixel vision sensors (DAVIS) \cite{brandli2014240,berner2013240}, which can simultaneously capture pixel-wise images and event data. Leveraging this advantage, we provide a novel perspective of training an event encoder using abundant paired event-image data instead of directly conducting event-text alignment, thus bypassing the shortage of large-scale paired event-text data. Instead of full finetuning the event encoder, we introduce a simple yet highly-efficient training strategy based on the LoRA \cite{hu2021lora} technique to focus on relating the event and image modalities, meanwhile preserving the highly-robust zero-shot ability provided by CLIP. In this way, CEIA can learn an event embedding space aligned with the image embedding space of frozen CLIP. Notably, the image space is already aligned with the text space during pretraining by CLIP. Consequently, event and text data are also naturally aligned by using image data as a bridge. By this way, CEIA can not only enhance open-world event-text understanding but also open the door to more event-based multi-modal understanding tasks \cite{xiao2022eva,lin2020learning,weng2023event,das2024halsie}.

In highlighting the versatility of CEIA, we make extensive evaluations through a diverse range of multi-modal understanding tasks. CEIA, designed to strike a unified embedding space for aligning event, image, and text data, can be smoothly applied to object recognition, event-image retrieval, event-text retrieval, and domain adaptation \cite{sun2022ess, messikommer2022bridging}. The experimental outcomes demonstrate that the state-of-the-art zero-shot performance can be achieved by CEIA over the existing methods, which further spotlights CEIA’s transferability and versatility. Additionally, we observe that, leveraging more training data, CEIA also exhibits the flexibility to yield a significant performance boost, ensuring the scalable capability. Through these extensive experimental evaluations on four applications, as shown in \cref{fig:ELIP}, we not only confirm the exceptional functionality of event, image, and text alignment of CEIA, but also underscore the comprehensive application capabilities of CEIA. We believe that CEIA stands as a robust and effective framework for open-world event-based multi-modal understanding.

In summary, CEIA presents three main contributions: (i) an effective framework to provide a novel perspective of learning to align event and image data as an alternative, thus bypassing the shortage of event-text datasets. (ii) a simple yet highly-efficient strategy for training the event encoder with the LoRA technique, meanwhile preserving the CLIP's powerful robustness. (iii) state-of-the-art results on four event-based multi-modal downstream tasks, including zero-shot and few-shot object recognition, event-text retrieval, event-image retrieval, and domain adaptation.

% In summary, CEIA presents three main contributions: (i) a novel framework to simultaneously align event, image and text modalities by only conducting event-image alignment, to overcome the shortage of paired event-text data, (ii) a simple yet highly-efficient strategy for training the event encoder with the LoRA technique, meanwhile preserving the highly-robust zero-shot capability, (iii) state-of-the-art results on four event-based multi-modal downstream tasks, including zero-shot and few-shot object recognition, event-text retrieval, event-image retrieval, and domain adaptation.

% In summary, our work presents three main contributions: (i) a scalable event-image pre-training method that overcomes the shortage of event-language data to bridge the domain gap, (ii) state-of-the-art zero-shot and few-shot results on three datasets, and (iii) extended support for various novel event-based cross-modal applications, including event-image retrieval, event-text retrieval, and DA.

\section{Related Work}
\subsection{Transferring CLIP}
In the image-based vision, pretrained Visual Language Models like CLIP \cite{radford2021learning}, ALIGN \cite{jia2021scaling}, and Florence \cite{yuan2021florence} demonstrate very impressive zero-shot transfer and generalization capabilities. Subsequently, a large number of follow-up works have been proposed to transfer the pretrained CLIP to more downstream tasks. For example, PointCLIP \cite{zhang2022pointclip} transforms 3D point clouds into a set of depth maps for zero-shot 3D object recognition, while DenseCLIP \cite{rao2022denseclip} converts the original image-text matching to pixel-text matching to guide the learning of dense prediction models. X-CLIP \cite{ni2022expanding} proposes a novel cross-frame attention mechanism to effectively expand CLIP to the video domain.

Recently, some works have applied Visual Language Models to event-based vision, demonstrating promising results. Two works closely related to ours are EventCLIP \cite{wu2023eventclip} and E-CLIP \cite{zhou2023clip}. Similar to PointCLIP, EventCLIP first transforms events into 2D frames and then uses frozen CLIP directly for zero-shot event object recognition. Following EventCLIP, E-CLIP focuses on advancing few-shot and standard object recognition. It introduces a novel event encoder for event temporal modeling and presents a triple contrastive alignment module to enable efficient knowledge transfer. In contrast, instead of directly utilizing frozen CLIP, we leverage existing abundant event-image datasets to adapt CLIP to event-based zero-shot tasks. 

\subsection{Multi-Modal Learning}
With the availability of large-scale multi-modal datasets, an increasing number of multi-modal foundation models have emerged. Some representative models are driving multi-modal learning, which has marked a significant advancement in AI evolution. For example, CLIP \cite{radford2021learning} demonstrates impressive zero-shot object recognition performance, while BLIP-2 \cite{li2023blip} exhibits capabilities approaching human-level performance in visual dialog, visual knowledge reasoning, and personalized image-to-text generation. Furthermore, Stable Diffusion \cite{rombach2022high} can generate realistic and accurate images based on given text conditions. InstructPix2Pix \cite{brooks2023instructpix2pix} can execute diverse image edits following human-written instructions, including object replacement, style modification, setting changes, and adjustments to the artistic medium.

These advancements motivate us to explore event-based multi-modal tasks. In this paper, we study two main directions. One is event-text understanding, including zero-shot learning and event-text retrieval, while the other is event-image understanding, involving event-image retrieval and domain adaptation. Our future work will focus on generalizing CEIA for wider multi-modal tasks, such as event-assisted video frame interpolation \cite{paikin2021efi,yu2021training,tulyakov2022time,xiao2022eva} and event-assisted motion deblurring \cite{lin2020learning,zhang2022unifying,weng2023event,zhou2023deblurring}.

\begin{figure}[tb]
  \centering
  \includegraphics[width=0.95\textwidth]{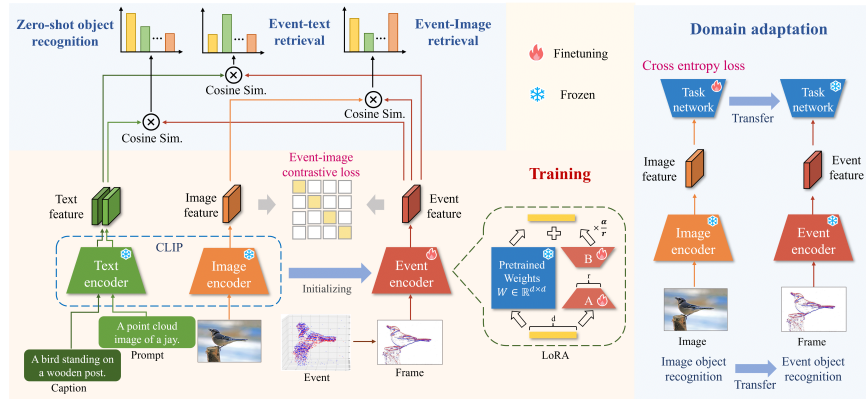}
  \caption{Overview of CEIA, which consists of a learnable event encoder, a frozen image encoder, and a frozen text encoder. We initialize the event encoder with CLIP's image encoder and finetune it using the LoRA \cite{hu2021lora} technique. We align the event embedding space and image embedding space through contrastive learning. In highlighting the versatility of CEIA, we make evaluations on four applications: object recognition, event-image retrieval, event-text retrieval, and domain adaptation.
  % We propose a straightforward yet efficient framework to adapt CLIP to event data. We provide a novel perspective of focusing on learning to align event and image data as an alternative, thus bypassing the shortage of event-text datasets.  In highlighting the versatility of our framework, we thoroughly evaluate CEIA on four applications: object recognition, event-image retrieval, event-text retrieval, and domain adaptation.
  }
  \label{fig:method}
  
\end{figure}

\section{Method}

% \subsection{Challenge and Motivation.}
% In particular, open-world event-based multi-modal understanding still remains under-explored. Our goal is to transfer the zero-shot capability of CLIP into event-based vision. To this end, two challenges need to be addressed. First, compared with natural images, the event data, captured by detecting the intensity changes, is essentially a kind of spatial-temporal data. Therefore, the big modality disparity makes it difficult to directly apply the image encoder of CLIP to event data. Second, intuitively, one way to achieve open-world event-based understanding is to train a large event-language model to accommodate \cref{equ:3} directly. Nevertheless, it is severely impeded due to the shortage of large-scale paired event-language datasets.

% Intuitively, one way to achieve open-world event-based understanding is to train a large event-language model. Nevertheless, it is severely impeded due to the shortage of paired event-language datasets. In contrast, CEIA provides a possible solution to make up the shortage of paired event-language data, which focuses on learning to align event and image data as an alternative. In this section, we first offer brief reviews of CLIP and EventCLIP (\cref{sec:reviews}). Following that, we show the technical details and fundamental principles behind our approach (\cref{sec:aligning}). 

% \subsection{Event-Text Alignment via Prompt Generation}

\subsection{CLIP Preliminaries}
\label{sec:reviews}
CLIP \cite{radford2021learning} is a visual-text pre-training method for image and text matching. Conceptually, CLIP consists of two encoders: an image encoder $\Phi_{image}(\cdot; \theta_{0})$ for extracting visual features and a text encoder $\Phi_{text}(\cdot; \theta_{1})$ for extracting text features. During training, CLIP utilizes 400 million training image-text pairs collected from the internet and employs a contrastive loss to learn a unified embedding space for accommodating image and text data. Specifically, given a set of image-text pairs $\left\{\mathbf{x}^{image}, \mathbf{x}^{text}\right\}$, CLIP is trained to search optimized parameters $\theta_{0}$ and $\theta_{1}$ to approach 
\begin{equation}
\label{equ:1}
\begin{aligned}
\Phi_{image}(\mathbf{x}^{image}; \theta_{0}) &= \Phi_{text}(\mathbf{x}^{text}; \theta_{1}). \\
\end{aligned}
\end{equation}
Note that we use ``$=$'' to denote the alignment in the whole paper. Leveraging the large-scale image-text dataset, CLIP demonstrates promising zero-shot performance for many downstream tasks, ensuring the incorporation of a huge range of visual concepts.
% Although the zero-shot inference does not require any new training data to fine-tune the model, it achieves promising classification performance.

% This allows CLIP to align images with any unseen semantic concepts in the open-set for zero-shot classification. Specifically, CLIP first constructs the text prompts by inserting all class names from the new dataset into handcrafted templates (e.g., ``a photo of a [CLASS]''). leveraging CLIP's remarkable visual-language alignment capability, the text features $W_{t}\in \mathbb{R}^{K\times C}$ encoded from the prompts can naturally function as the weights of the zero-shot visual classifier. Meanwhile, CLIP's visual encoder extracts visual features $F_{v}\in \mathbb{R}^{1\times C}$ from the images. The predicted probabilities $p\in \mathbb{R}^{1\times K}$ are computed via the classifier as follows:

% \begin{equation}
%   logits=F_{v}W_{t}^{T};\quad p=softmax(logits),
%   \label{eq:zeroshot}
% \end{equation}

% Although the zero-shot inference does not require any new training data to fine-tune the model, it achieves promising classification performance.

\subsection{The CEIA Framework}
In particular, open-world event-based multi-modal understanding still remains under-explored. Our goal is to transfer the zero-shot capability of CLIP into the event-based vision. To this end, two challenges need to be addressed. First, intuitively, one way to achieve open-world event-based understanding is to train a large event-text model. Nevertheless, it is severely impeded due to the shortage of large-scale paired event-text data. Second, compared with natural images, the event data, captured by detecting the intensity changes, is essentially a kind of spatial-temporal data. Therefore, the big modality disparity makes it difficult to directly apply the image encoder of CLIP to event data. 

In response to these two challenges, CEIA makes two key modifications. First, CEIA provides a novel perspective of focusing on learning to align event and image data instead of conducting event-text alignment, thus bypassing the shortage of large-scale paired event-text data. Second, CEIA learns an individual event encoder to alleviate the event-image modality disparity instead of directly utilizing the frozen image encoder like EventCLIP \cite{wu2023eventclip}. In the following, we will formally introduce the method.

\noindent \textbf{Overview.} \cref{fig:method} shows an overview of CEIA, which is composed of a frozen image encoder $\Phi_{image}(\cdot; \theta_{0})$, a frozen text encoder $\Phi_{text}(\cdot; \theta_{1})$ and a learnable event encoder $\Phi_{event}(\cdot; \theta_{2})$. Given a triple set of image-event-text pairs $\left\{\mathbf{x}^{\text{event}}, \mathbf{x}^{\text{image}}, \right.$\\
$\left. \mathbf{x}^{\text{text}}\right\}$, CEIA learns to search a desirable parameter $\theta_{2}$, which meets the following requirement:
\begin{equation}
\label{equ:2}
\begin{aligned}
\Phi_{event}(\mathbf{x}^{event}; \theta_{2}) &= \Phi_{image}(\mathbf{x}^{image}; \theta_{0})
\end{aligned}
\end{equation}
Notably, CLIP has already provided the powerful image-text alignment as shown in \cref{equ:1}.
% Recalling the image-language alignment provided by CLIP, we have the following equation:
% \begin{equation}
% \label{equ:2}
% \begin{aligned}
% \Phi_{image}(\mathbf{x}_{i}^{image}; \theta_{1}) &= \Phi_{text}(\mathbf{x}_{i}^{text}; \theta_{2}). \\
% \end{aligned}
% \end{equation}
Consequently, through combining \cref{equ:1} and \cref{equ:2}, we can align event and text data by regarding $\Phi_{image}(\mathbf{x}^{image}; \theta_{0})$ as a bridge
\begin{equation}
\label{equ:3}
\begin{aligned}
\Phi_{event}(\mathbf{x}^{event}; \theta_{2}) &= \Phi_{text}(\mathbf{x}^{text}; \theta_{1}). \\
\end{aligned}
\end{equation}

\noindent \textbf{Event Encoder.} 
Following existing research \cite{zhou2023clip,yang2023event,klenk2024masked}, we selected the Vision Transformer \cite{dosovitskiy2020image}, a reliable and widely-used model, as our event encoder. Leveraging the unified encoder architecture, we propose initializing the event encoder with CLIP’s image encoder and then finetuning it, instead of training it from scratch. This initialization transfers spatial prior knowledge from images to events, accelerating the training process and enhancing the data efficiency of CEIA. In our experiments, this initialization proved not only beneficial but also essential. Since the training data is still too limited for cross-modal alignment, we attempted to train the event encoder from scratch, but failed.

% Even with relatively small-scale training data, CEIA has already achieved competitive zero-shot performance. 
\noindent \textbf{Event Representations.}
We explored various event representations and determined that the red-blue color map, commonly used for visualizing events, is the most effective. This choice minimizes the difference between the event representation and the natural images used by CLIP, thereby simplifying cross-modal alignment.

\noindent \textbf{LoRA-Based Finetuning.}
Intuitively, one simple way to learn an event encoder is full finetuning. However, it will destroy the original CLIP’s weights, which brings the inferior zero-shot capability. Recently, LoRA \cite{hu2021lora} stands out as one of the best parameter-efficient transfer learning methods, which has been widely adopted to finetune many LLMs. Specifically, LoRA \cite{hu2021lora} shows that the pretrained models can still learn efficiently even when projected into a smaller subspace. For each pretrained weight matrix $W_{0}\in \mathbb{R}^{d\times k}$, we can replace its update with a low-rank decomposition $\Delta W=BA$, where $B\in \mathbb{R}^{d\times r}$, $A\in \mathbb{R}^{r\times k}$. Note that $W_{0}$ is frozen, while $A$ and $B$ are trainable. For the original forward pass $h=W_{0}x$, the modified forward pass is:
\begin{equation}
h=W_{0}x+\Delta Wx=W_{0}x+\frac{\alpha}{r} BAx
\end{equation}
where $\alpha$ is a hyperparameter used to adjust the influence of the new parameters. LoRA-based finetuning provides three key advantages for CEIA: 1) It avoids catastrophic forgetting, thus preserving CLIP's strong generalization and zero-shot capabilities. 2) It prevents overfitting to the limited training data. 3) It significantly reduces training time and memory costs.

\noindent \textbf{Event-Image Contrastive Learning.} 
The objective of training our event encoder $\Phi_{event}(\cdot; \theta_{2})$ is to minimize the distance between the frames transformed from events and images in the same pair, while maximizing the distance of others. We draw the inspiration from many methods \cite{radford2021learning,xue2023ulip,hegde2023clip,zhou2023clip,guzhov2022audioclip}, which advocates the utilization of multi-modal contrastive learning.
Specifically, given a set of event-image pairs $\left\{\mathbf{x}_{i}^{event},\mathbf{x}_{i}^{image}\right\}_{i=1}^{N}$, 
% As described in \cref{sec:intro}, we can leverage abundant paired event-image data as an alternative to overcome the lack of available paired event-language data. For a set of event-image pairs $\left\{x_{i}^{event},x_{i}^{image}\right\}_{i=1}^{N}$, cross-modal contrastive learning aims to minimize the distance between the frames transformed from events and images in the same pair, while maximizing the distance of others. Specifically, 
we encode them into normalized embeddings: $f_{i}^{event}=\Phi_{event}(\mathbf{x}_{i}^{event}; \theta_{2})$ and $f_{i}^{image}=\Phi_{image}(\mathbf{x}_{i}^{image}, \theta_{0})$. 
Denoting $M_{1}$ and $M_{2}$ are two modalities, the InfoNCE \cite{oord2018representation} loss can be formulated as
\begin{equation}
  % L(M_{1},M_{2})=-log\frac{exp(f_{i}^{M_{1}}\cdot f_{i}^{M_{2}}/\tau)}{exp(f_{i}^{M_{1}}\cdot f_{i}^{M_{2}}/\tau)+ {\textstyle \sum_{j\ne i}} exp(f_{i}^{M_{1}}\cdot f_{j}^{M_{2}}/\tau)},
L(M_{1},M_{2})=-log\frac{exp(f_{i}^{M_{1}}\cdot \left ( f_{i}^{M_{2}} \right ) ^{T} /\tau)}{exp(f_{i}^{M_{1}}\cdot \left ( f_{i}^{M_{2}} \right ) ^{T}/\tau)+ {\textstyle \sum_{j\ne i}} exp(f_{i}^{M_{1}}\cdot \left ( f_{j}^{M_{2}} \right ) ^{T}/\tau)},
  \label{eq:contrastive_loss}
\end{equation}
where $\tau$ is a learnable temperature parameter to control the smoothness of the softmax distribution. 
Following CLIP \cite{radford2021learning}, we consider every example $j\ne i$ in the mini-batch as a negative. Finally, the weights of the event encoder $\theta_{2}$ is optimized by minimizing a symmetric InfoNCE loss
\begin{equation}
  L_{final}=L(event,image)+L(image,event).
  \label{eq:final_loss}
\end{equation}

Through event-image contrastive learning, we can align representations of event, image, and text modalities into the same embedding space. 
In the following, we will elaborate on the details about how to extend CEIA to open-world event-based multi-modal applications.

\subsection{Event-Based Multi-Modal Applications}

\noindent \textbf{Object Recognition.} 
Zero-shot object recognition aims to classify objects that are not included in the training dataset. As shown in \cref{equ:3}, CEIA has achieved event-text alignment in an indirect manner. Through this event-text alignment, CEIA enables zero-shot event-based object recognition. Specifically, we first construct text prompts by inserting the class names of new objects into predefined templates (e.g., ``image of a [CLASS]''). Then, we extract their textual features $W_{t}$ by $\Phi_{text}(\cdot; \theta_{2})$. Since each row vector in $W_{t}$ encodes class knowledge, $W_{t}$ can naturally function as the zero-shot event classifier. Meanwhile, we utilize $\Phi_{event}(\cdot; \theta_{1})$ to extract the event features $f_{i}^{event}$ from the input events. Finally, the predicted probabilities for K classes are computed via the classifier as follows:

\begin{equation}
logits_{i}=f_{i}^{event}W_{t}^{T};p_{i}=softmax(logits_{i}).
  \label{eq:zeroshot}
\end{equation}
Similarly, \cref{eq:zeroshot} can be also utilized for few-shot object recognition.

\noindent \textbf{Event-Image/Event-Text Retrieval.} 
Event-image retrieval refers to the task of searching for the most related image in a large-scale image dataset based on a given event, or vice versa. For instance, when given an image query $\mathbf{x}^{image}_{q}$, we first extract its image feature $f^{image}_{q}$ using $\Phi_{image}(\cdot; \theta_{0})$. Then, we feed forward all event examples $\left\{\mathbf{x}^{event}_{j}\right\}_{j=1}^{N}$ into $\Phi_{event}(\cdot; \theta_{2})$ to obtain $\left\{f^{event}_{j}\right\}_{j=1}^{N}$. Subsequently, we calculate their cosine similarity and retrieve the most related event $\mathbf{x}^{event}_{j^{\ast}}$ with the highest similarity score:

\begin{equation}
\begin{aligned}
% j^{\ast}=\underset{j}{argmax}\left(\frac{f_{q}^{image} \left ( f_{j}^{event} \right )^{T}  }{\left \| f_{q}^{image} \right \| \left \| f_{j}^{event} \right \|}\right) \\
j^{\ast}=\underset{j}{\text{argmax}}\left(\frac{\displaystyle f_{q}^{\text{image}} \left ( f_{j}^{\text{event}} \right )^{T}  }{\displaystyle \left \| f_{q}^{\text{image}} \right \| \left \| f_{j}^{\text{event}} \right \|}\right)
  \label{eq:ei_retrieval}
\end{aligned}
\end{equation}
For event-text retrieval, we calculate the similarity score between event features and text features and select the item with the highest similarity score.
% Similarly, we employ the same method for event-text retrieval.

\noindent \textbf{Domain Adaptation.} Domain adaptation \cite{farahani2021brief,sun2022ess,messikommer2022bridging} aims to transfer tasks from a labeled source domain (images) to a target domain (events). It can leverage existing image datasets to train models, thereby overcoming the lack of high-quality labeled event datasets. Specifically, As depicted in \cref{fig:method}, we conducted domain adaptation experiments on object recognition. Formally, denote $f^{image}$ and $l$ as the image feature extracted by $\Phi_{image}(\cdot; \theta_{0})$ and the available label, respectively. We train a task network $T(\cdot;\theta_{4})$, whose weights $\theta_{4}$ are optimized by minimizing the commonly-used soft-max cross-entropy loss:

\begin{equation}
logits=T(f^{image};\theta_{4});L_{image} =CrossEntropy(logits,l).
  \label{eq:uda_image}
\end{equation}
Subsequently, we directly apply the trained task network $T(\cdot;\theta_{4})$ to the event domain to generate predictions:
\begin{equation}
pred=T(f^{event};\theta_{4}).
  \label{eq:uda_event}
\end{equation}
As shown in \cref{equ:2}, CEIA has already aligned event and image data, which ensures the transferability and applicability of the network $T(\cdot;\theta_{4})$ when applied for event data.

\section{Experiments}

\subsection{Dataset Preparation}
\label{sec:pretrain_datasets}
% We use the following four public datasets to evaluate four multi-modal downstream tasks.

\noindent\textbf{N-ImageNet.} 
N-ImageNet \cite{kim2021n} is built by moving an event camera in front of an LCD monitor which displays images from ImageNet \cite{deng2009imagenet}.
We leverage the event-image pairs from N-ImageNet \cite{kim2021n} and ImageNet-1K \cite{deng2009imagenet} for training. 
Similar to ImageNet-1K, N-ImageNet contains 1.78 million event streams belonging to 1,000 classes. For training, we split N-ImageNet to construct two subset datasets: the Small dataset includes 129,393 event streams belonging to the first 100 classes and the Large dataset includes 638,878 belonging to the first 500 classes.
We call the method ``X'' trained on Small and Large datasets as ``X-S'' and ``X-L'', respectively.
We use the Small and Large datasets to explore the scalable capability of our approach. 
We utilize the official splitting to obtain the training and test datasets.

% we respectively utilize event-image pairs from only the first 100 classes (called \textbf{S}) and the first 500 classes (called \textbf{L}) for training. This confirms the scalability of our approach while reserving the classes from 500 to 1,000 for downstream tasks. 
% To demonstrate the robustness of our approach, we also employ three additional out-of-distribution datasets \textit{only for evaluations}.
% \subsection{Downstream Datasets}
% We use the following four public datasets for both event-based zero-shot and few-shot classification:

% \subsubsection{N-ImageNet.} As described in \cref{sec:pretrain_datasets}, for downstream tasks, we utilize event-image pairs from classes 500 to 1,000 of both N-ImageNet and ImageNet-1K. To demonstrate the robustness of our approach, we also employ three additional out-of-distribution datasets.

\noindent\textbf{N-Caltech101.} 
Similar to N-ImageNet \cite{kim2021n}, N-Caltech101 \cite{orchard2015converting} is built by moving a 180$\times$240 resolution ATIS event camera in front of a monitor displaying still images from Caltech101 \cite{fei2004learning}. 
It contains 8,246 samples, each with a duration of 300 ms, belonging to 101 classes. 
We adopt the same splitting strategy as EST \cite{gehrig2019end} to obtain the training and test datasets.

\noindent\textbf{CIFAR10-DVS.} 
Unlike N-Caltech101 \cite{orchard2015converting} and N-ImageNet \cite{kim2021n}, CIFAR10-DVS \cite{li2017cifar10} is created through repeating smooth movements of images on an LCD monitor in front of a DVS camera. This process converts the popular CIFAR-10 \cite{krizhevsky2009learning} dataset into 10,000 event streams across 10 different classes. 
We randomly allocate 4,000 samples for the test set and 6,000 samples for the training set.

\noindent\textbf{ASL-DVS.} 
ASL-DVS \cite{bi2020graph} is a relatively complex dataset containing the second largest number of labeled examples. It contains 24 classes corresponding to 24 letters (A-Y, excluding J) of the American Sign Language. For each letter, 4,200 samples are collected by capturing real-world events. Each sample spans approximately 100 milliseconds. 
We randomly select 1,000 samples for the test set and 3,200 samples for the training set.

\noindent\textbf{NIN-Prompt/NIN-BLIP2/NIN-BLIP2-retrieval.} 
Considering the shortage of currently available large-scale event-text datasets, we make the first attempt to build two kinds of event-text datasets based on N-ImageNet for training, denoted as ``NIN-Prompt'' and ``NIN-BLIP2''. Specifically, for ``NIN-Prompt'', we first create prompts by placing the class names of events into the template ``A point cloud image of [CLASS]''. We then use these prompts as captions for corresponding events. For ``NIN-BLIP2'', we utilize BLIP2 \cite{li2023blip} with the frozen LLM OPT \cite{zhang2022opt} to conduct zero-shot image captioning, generating high-quality captions for the images from ImageNet. Subsequently, we pair them with events from N-ImageNet to construct the dataset.
Furthermore, we create a test dataset, named ``NIN-BLIP2-retrieval'', for evaluating event-text retrieval. However, the captions generated from images belonging to the same class are too similar, which may correspond to multiple events. To mitigate this issue, we selectively sample only five images from each class to generate captions, constructing a test set containing 2,500 event-text pairs.

% Considering the shortage of currently available large-scale event-text dataset, we make the first attempt to build two kinds of event-text datasets based on N-ImageNet, denoted as ``N-ImageNet-Prompt'' and ``N-ImageNet-BLIP2''.
% Specifically, ``N-ImageNet-Prompt'' is built by directly converting class name to a prompt with the template like ``A point cloud image of [CLASS]''.
% For ``N-ImageNet-BLIP2'', we employed BLIP2 \cite{li2023blip} with the frozen LLM OPT \cite{zhang2022opt} to conduct zero-shot image captioning, generating high-quality captions for the images from ImageNet.
% To ensure the generated captions distinguishable, we selectively sample only five images from each class (from classes 500 to 1,000).
% Note that these two datasets are only used for training.
% Notice that, for N-ImageNet, we utilize the official splitting to obtain the training and test datasets. For N-Caltech, we adopt the same splitting as EST \cite{gehrig2019end}. However, CIFAR10-DVS and ASL-DVS do not provide official splitting. Consequently, we randomly split the datasets. Specifically, for CIFAR10-DVS, we randomly allocated 4,000 samples for the test set and 6,000 samples for the training set. As for ASL-DVS, we randomly selected 1,000 samples for the test set and 3,200 samples for the training set.

% Note that CIFAR10-DVS and ASL-DVS have not provided predefined training and testing splits, so we randomly split the dataset.

\subsection{Implementation Details}
We initialize our event encoder with the ViT-L/14 \cite{dosovitskiy2020image} image encoder of CLIP. The AdamW \cite{loshchilov2017decoupled} optimizer and a cosine schedule warm-up learning rate schedule \cite{loshchilov2016sgdr} are adopted for training. 
For LoRA-based finetuning \cite{hu2021lora}, we set the peak learning rate to $5\times 10^{-4}$ and the weight decay to $1\times 10^{-2}$. 
For full finetuning, we set the peak learning rate to $1\times 10^{-7}$ and the weight decay to $1\times 10^{-1}$. 
The training batch size is set to 128 for all experiments. Additionally, we conduct prompt engineering and create task-relevant templates for each dataset. 
Specifically, we adopt ``A point cloud image representing the American Sign Language letter [CLASS]'' for ASL-DVS, ``Image of a [CLASS]'' for N-Caltech101, and ``A point cloud image of a [CLASS]'' for CIFAR10-DVS and N-ImageNet.

\subsection{Baselines}
% Event-based zero-shot classification is very challenging because models lack visibility of the classes in the test set, making it difficult to adapt to new data distributions.
% As far as we know, EventCLIP and E-CLIP are the only two attempts at event-based zero-shot classification. Among them, E-CLIP introduces a novel Hierarchical Triple Contrastive Alignment module aimed at enhancing few-shot and standard classification performance. However, it paradoxically results in poorer zero-shot performance. As a result, EventCLIP maintains its status as the state-of-the-art for zero-shot classification. Therefore, we use EventCLIP as our major baseline.
We compare CEIA with the current state-of-the-art event-based zero-shot method, EventCLIP \cite{wu2023eventclip}. 
Additionally, we combine the pre-trained event-based video reconstruction network E2VID \cite{rebecq2019events} with the frozen CLIP to construct another simple zero-shot method, which is denoted as ``E2VID-CLIP''. 

Moreover, leveraging our building event-text datasets NIN-Prompt and NIN-BLIP2, we are able to directly train a CLIP-based event-text alignment model, called ``CETA''.
We denote ``CETA'' trained on such two datasets as ``CETA-Prompt'' and ``CETA-BLIP2'', respectively.

% Furthermore, we wonder to how competitive CEIA would remain even after the emergence of the first event-language dataset in the future. 
% Therefore, we introduce two baselines for generating paired captions to train a event-language model: 1) Directly using prompts created from class names as paired captions (called CETA-Prompt); 2) Utilizing BLIP2\cite{li2023blip}, known for its robust zero-shot image-to-text generation capabilities, to generate paired captions (called CETA-BLIP2). 

\subsection{Object Recognition}
\begin{table}[tb]
  \caption{Quantitative results of zero-shot object recognition.}
  \vspace{-3mm}
  \label{tab:zero_shot}
  \centering    
  \setlength{\tabcolsep}{4pt}
  \renewcommand{\arraystretch}{1}
  \resizebox{1.0\linewidth}{!}{
  \begin{tabular}{c|cc|cc|cc|cc}
    \toprule

    \multirow{3}{*}{\centering Method}
    & \multicolumn{2}{c|}{In-Distribution} & \multicolumn{6}{c}{Out-of-Distribution}\\
    \cmidrule(lr){2-3} \cmidrule(lr){4-9}
        
    & \multicolumn{2}{c|}{N-ImageNet \cite{kim2021n}} & \multicolumn{2}{c|}{N-Caltech101 \cite{orchard2015converting}}   & \multicolumn{2}{c|}{CIFAR10-DVS \cite{li2017cifar10}} & \multicolumn{2}{c}{ASL-DVS \cite{bi2020graph}}\\
    
    \cmidrule(lr){2-3} \cmidrule(lr){4-5} \cmidrule(lr){6-7} \cmidrule(lr){8-9}
                       & Acc1 & Acc5 & Acc1 & Acc5 & Acc1 & Acc5 & Acc1 & Acc5\\
    \midrule
    
    EventCLIP \cite{wu2023eventclip}        & 26.72 & 39.39 & 69.73 & 85.93     
                      & 13.23 & 56.17 & 8.72  & 25.97 \\
                      
    E2VID-CLIP        & 13.68 & 27.44 & \textbf{82.53} & \textbf{93.62} 
                      & 13.85 & 55.57 & 8.43  & 25.82 \\
                      
    CETA-Prompt-S & 29.35 & 50.73 & 70.93 & 86.61 
                      & 14.24 & 59.35 & 8.87  & 28.23 \\
                   
    CETA-BLIP2-S        & 33.86 & 57.53 & 75.01 & 88.68 
                      & 16.27 & 65.49 & 7.53  & 24.88 \\
    \midrule
    
    % ELIP-Full (S)& 35.14 & 58.17 & 75.24 & 88.62
                    % & 17.24 & 66.27 & 9.29  & 29.30 \\
                    
    CEIA-S  & \underline{37.25} & \underline{61.60} & 72.31 & 86.50
                      & \underline{18.40} & \underline{65.75} & \underline{12.11} & \textbf{32.62} \\
                      
    % ELIP-Full (L)& 42.44 & 67.36 & 80.41 & 91.15 
                    % & 20.84 & 71.05 & 10.50 & 30.30  \\
                    
    CEIA-L  & \textbf{43.68} & \textbf{68.78} & \underline{79.20} & \underline{90.80}
                      & \textbf{22.20} & \textbf{69.07} & \textbf{12.67} & \underline{31.20} \\
    
    \bottomrule
  \end{tabular}
  }
  % \vspace{-6mm}
\end{table}

\noindent \textbf{Metrics.}
We evaluate the performance of object recognition in terms of the common top-1 accuracy (Acc1) and top-5 accuracy (Acc5) \cite{he2016deep,simonyan2014very}. 

\noindent \textbf{Zero-Shot Results.}
Event-based zero-shot object recognition is a challenging task because the classes in the test set are unseen to the model during training. 
We report the in-distribution and out-of-distribution results in \cref{tab:zero_shot}.
% We report the zero-shot object recognition results of our CEIA training on N-ImageNet (S) and N-ImageNet (L) in \cref{tab:zero_shot}. 
The experimental results indicate that our CEIA consistently outperforms the state-of-the-art baselines across all datasets.  
For instance, on N-ImageNet and N-Caltech101, CEIA-L achieves improvements of 16.96\% and 9.47\% in top-1 accuracy compared with EventCLIP, respectively. 
These improvements highlight the effectiveness of our CEIA for open-world event-based understanding. 
Although E2VID-CLIP achieves better results than ours on N-Caltech101, its complex reconstruction network introduces significant inference latency.
% Moreover, we observed relatively lower improvements in ASL-DVS. This could be attributed to the gap between the training dataset N-ImageNet, generated by displaying images on a monitor, and ASL-DVS, built by capturing real-world events. 

Besides, we notice that CETA-BLIP2 achieves better zero-shot results than CETA-Prompt, which can be attributed to the reason that BLIP2 is able to generate more accurate captions compared with the simple prompt template. However, the event-text alignment method CETA-BLIP2 (CETA-Prompt) exhibits inferior results compared with our event-image alignment method CEIA, highlighting the effectiveness of our event-image alignment strategy compared with the direct event-text alignment strategy.

% On N-ImageNet, N-Caltech, and CIFAR10-DVS, CETA-BLIP2 (S) significantly outperforms CETA-Prompt (S), due to its captions being more detailed and natural. However, our CEIA (S) trained using event-image contrastive learning achieves better performance. This could be because the frame-like representation of events is more easily aligned with natural images compared to the textual modality.

% Furthermore, as shown in \cref{fig:scale}, we can see that CEIA (L), which is pretrained on the larger-scale event-image pairs, achieves significantly better performance across all benchmarks. This indicates that, leveraging more training data, CEIA can exhibit the flexibility to boost performance, ensuring its scalable capability. Therefore, larger-scale event-image pretraining is an exciting direction for future work.

% Additionally, due to its ability to avoid overfitting to the pre-training dataset, ELIP-LoRA (L) outperforms ELIP-Full (L) in top-1 accuracy across all benchmarks except N-Caltech. However, we speculate that the poorer performance on N-Caltech may attributed to too few trainable parameters, leading to underfitting.

% \subsection{Few-Shot object recognition}

\begin{table}[tb]
  \caption{Quantitative results of few-shot object recognition. Acc1 (\%) is reported.}
  \vspace{-3mm}
  \label{tab:few_shot}
  \centering
  \setlength{\tabcolsep}{1pt}
  \renewcommand{\arraystretch}{1}
  \resizebox{1.0\linewidth}{!}{
  \begin{tabular}{c|ccccc|ccccc}
    \toprule
    
    Datasets & \multicolumn{5}{c|}{N-ImageNet \cite{kim2021n}} & \multicolumn{5}{c}{N-Caltech101 \cite{orchard2015converting}}\\
    \midrule
    Data per Class  &   1    &    2   &   5    &   10   &   20   
                    &   1    &    2   &   5    &   10   &   20   \\
    \midrule
    Sorted Time Surface \cite{alzugaray2018ace}
                    & 1.24   & 2.19   & 4.26   & 7.53    & 12.81  
                    & 27.80  & 31.99  & 54.85  & 65.94   & 75.47 \\
                    
    DiST  \cite{kim2021n}          & 1.16   & 1.75   & 4.22   & 7.65    & 13.07     
                    & 26.42  & 28.14  & 53.48  & 65.71   & 73.92 \\
    
    EventCLIP \cite{wu2023eventclip}      & \underline{29.41}  & \underline{31.14}  & \underline{32.56}  & \underline{33.08}   & \underline{36.40}   
                    & \underline{75.82}  & \underline{78.86}  & \underline{83.57}  & \underline{87.42}   & \underline{90.41} \\
    % E-CLIP          & -      & -      & -      & -       & -      
                    % & 66.72  & 75.87  & 82.35  & 86.92   & 90.51 \\
    % ELIP-Full(L) & 43.86  & 44.55  & 48.00  & 49.91   & -      
    %                 & 81.94  & 83.08  & 87.79  & 89.45   & 91.17 \\
    CEIA-L     & \textbf{44.77}  & \textbf{46.07}  & \textbf{49.58}  & \textbf{51.40}   & \textbf{53.32}  
                    & \textbf{84.46}  & \textbf{87.16}  & \textbf{89.28}  & \textbf{90.71}   & \textbf{92.14} \\
    % \midrule
    % Datasets & \multicolumn{5}{c|}{CIFAR10-DVS} & \multicolumn{5}{c}{ASL-DVS}\\
    % \midrule
    % Data per Class  &   1    &    2   &   5    &   10   &   20   
    %                 &   1    &    2   &   5    &   10   &   20   \\
    % \midrule
    % Sorted TimeSurface
    %                 & 16.50  & 20.10  & 24.80  & 30.58   & 38.40  
    %                 & 30.91  & 44.74  & 65.72  & \underline{77.55}   & \underline{90.01} \\
                    
    % DiST            & 16.05  & 21.03  & 21.42  & 29.30   & 33.63  
    %                 & \underline{36.72}  & \underline{49.42}  & \underline{67.32}  & 69.65   & \textbf{92.49} \\
    % EventCLIP       & \underline{21.28}  & \underline{26.95}  & \underline{29.90}  & \underline{35.98}   & \underline{41.93} 
    %                 & 27.04  & 41.15  & 58.63  & 68.20   & 79.56 \\
    % % E-CLIP          & -      & -      & -      & -       & -      
    % %                 & -      & -      & -      & -       & -     \\
    % % ELIP-Full(L) & 26.55  & 28.98  & 34.85  & 41.48   & 49.80  
    % %                 & 36.19  & 41.46  & 64.92  & 73.55   & 84.87 \\
    % CEIA (L) & \textbf{28.73}  & \textbf{31.75}  & \textbf{39.35}  & \textbf{49.30}   & \textbf{60.60}  
    %                 & \textbf{39.69} & \textbf{50.48}  & \textbf{69.61}  & \textbf{77.60}  & 86.95 \\
    
    \bottomrule
  \end{tabular}
  }
  \vspace{-6mm}
  
\end{table}

% \begin{figure}[tb]
%   \centering
%   \includegraphics[height=4cm]{repr.png}
%   \caption{Visualization of various representations.}
%   \label{fig:repr}
% \end{figure}
\noindent \textbf{Few-Shot Results.} 
We consider a general N-shot setting, i.e., N examples are randomly sampled from each class for training. We compare our CEIA with the current state-of-the-art few-shot classifier, EventCLIP \cite{wu2023eventclip}. In addition, we also compare with some representative methods without CLIP, namely, Sorted Time Surface \cite{alzugaray2018ace} and DiST {\cite{kim2021n}. Notice that, we follow the original papers and use ResNet34 \cite{he2016deep} pre-trained on ImageNet \cite{deng2009imagenet} as their backbone.

As shown in \cref{tab:few_shot}, in the extreme case of 1-shot, the performance of Sorted Time Surface and DiST drops significantly due to the serious lack of training data. In contrast, our CEIA can leverage CLIP's outstanding robustness to quickly adapt to the new distribution, demonstrating a large margin performance boost.
% Consequently, CEIA outperforms Sorted Time Surface and DiST by a very large margin. 
In terms of Acc1, CEIA-L outperforms Sorted Time Surface by 43.53\% and 56.66\% in terms of Acc1 on N-ImageNet and N-Caltech101 with 1-shot, respectively. 
Compared to EventCLIP, our CEIA-L achieves superior results on all datasets and all N-shot settings. This indicates that CEIA significantly enhances the transferability of knowledge from CLIP to event-based vision.
% Additionally, we also observed that the performance gap between CEIA and Sorted Time Surface and DiST decreases in ASL-DVS. This is attributed to the concept of letters from American Sign Language being too fine-grained for CLIP, thereby limiting its robustness to the new distribution.

% \section{Downstream Event-Based Cross-modal Applications}

\subsection{Event-Image Retrieval}

\noindent \textbf{Metrics.} 
We measure the performance of event-image retrieval through computing recall at K (R@K) \cite{lee2018stacked}, which is defined as the fraction of queries for which the correct item is retrieved in the closest K points to the query.

\begin{figure}[t!]
  \centering
  \includegraphics[width=12.1cm]{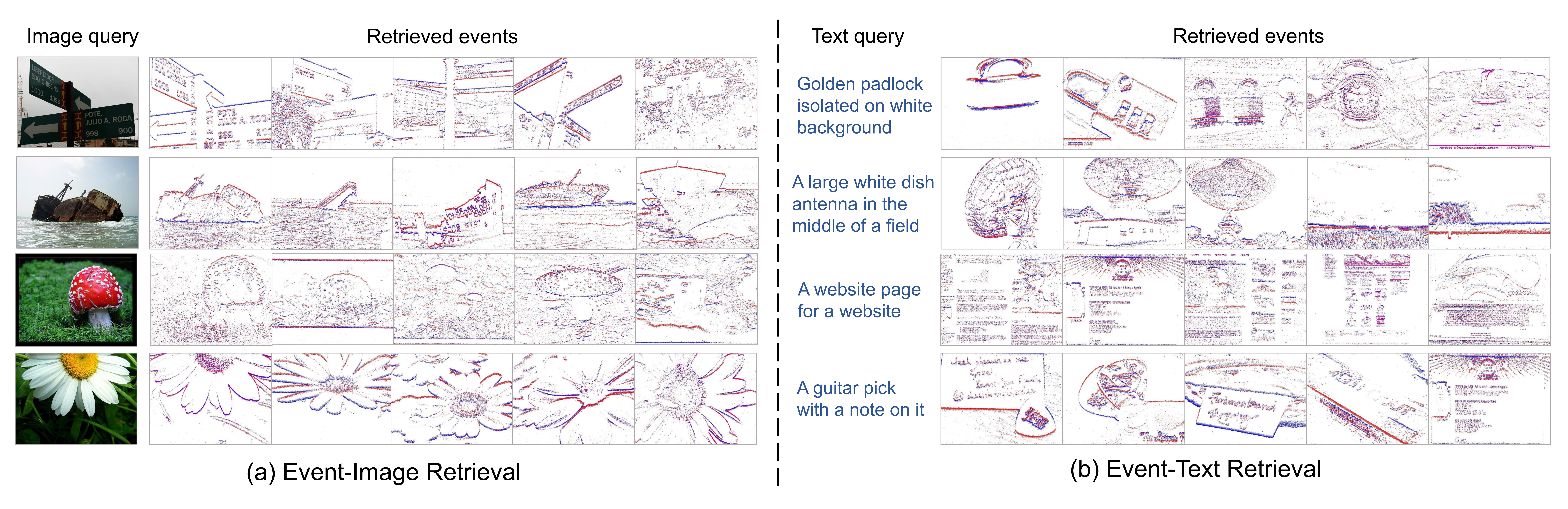}
  \vspace{-4mm}
  \caption{Qualitative results of event-image retrieval and event-text retrieval.}
  \label{fig:visualization}
\end{figure}

\begin{table}[t!]
  \caption{Quantitative results of event-image retrieval.}
  \vspace{-2mm}
  \label{tab:e_i_retrieval}
  \centering
  \setlength{\tabcolsep}{1pt}
  \renewcommand{\arraystretch}{1}
  \resizebox{1.0\linewidth}{!}{
  \begin{tabular}{c|ccc|ccc|ccc|ccc}
    \toprule

    \multirow{3}{*}{\centering Method}
    & \multicolumn{6}{c|}{N-ImageNet\cite{kim2021n}} & \multicolumn{6}{c}{N-Caltech101\cite{orchard2015converting}}\\
    \cmidrule(lr){2-7} \cmidrule(lr){8-13}
    & \multicolumn{3}{c|}{Event query} & \multicolumn{3}{c|}{Image query} & \multicolumn{3}{c|}{Event query} & \multicolumn{3}{c}{Image query}\\

    \cmidrule(lr){2-4} \cmidrule(lr){5-7} \cmidrule(lr){8-10} \cmidrule(lr){11-13}
    & R@1 & R@5 & R@10 & R@1 & R@5 & R@10 & R@1 & R@5 & R@10 & R@1 & R@5 & R@10\\

    \midrule
    EventCLIP \cite{wu2023eventclip}        & 3.61  &  7.67 & 10.74 & 11.33 & 20.29 & 25.74
                      & 9.53  & 24.46 & 33.54 & 11.96 & 23.77 & 28.47 \\
    E2VID-CLIP        & 2.49  &  4.38 & 5.43  &  5.79 & 11.40 & 14.62
                      & 23.24 & 50.02 & 63.69 & \underline{35.69} & \underline{60.32}  & \underline{72.90} \\
    CETA-Prompt-S  & 2.43  &  5.73 &  7.91 & 11.91 & 23.53 & 29.85
                      & 7.58  & 23.24 & 31.76 & 16.14 & 38.94 & 52.21 \\
    CETA-BLIP2-S   & 4.82  & 10.08 & 13.75 & 15.61 & 29.08 & 35.74
                      & 10.54 & 27.13 & 36.34 & 19.26 & 42.83 & 55.78\\

    \midrule
    
    CEIA-S       & \underline{35.24} & \underline{54.33} & \underline{61.73} & \underline{36.62} & \underline{54.73} & \underline{62.72}
                    & \underline{28.19} & \underline{52.98} & \underline{64.94} & 33.63 & 60.12 & 71.68 \\
    CEIA-L       & \textbf{41.22} & \textbf{61.39} & \textbf{69.61} & \textbf{44.15} & \textbf{62.86} & \textbf{70.12}
                    & \textbf{32.94} & \textbf{60.56} & \textbf{72.45} & \textbf{39.83} & \textbf{66.36} & \textbf{78.09}\\
    
    \bottomrule
  \end{tabular}
  }
  \vspace{-6mm}
  
\end{table}

\begin{table}[t!]
  \caption{Quantitative results of event-text retrieval on our built event-text dataset NIN-BLIP2-retrieval.}
  \vspace{-2mm}
  \label{tab:e_t_retrieval}
  \centering
  \setlength{\tabcolsep}{5pt}
  \renewcommand{\arraystretch}{1}
  \resizebox{0.8\linewidth}{!}{
  \begin{tabular}{c|ccc|ccc}
    \toprule

    \multirow{2}{*}{\centering Method}
    & \multicolumn{3}{c|}{Event query} & \multicolumn{3}{c}{Text query}\\

    \cmidrule(lr){2-4} \cmidrule(lr){5-7}
    & R@1 & R@5 & R@10 & R@1 & R@5 & R@10\\

    \midrule
    EventCLIP \cite{wu2023eventclip}        & 15.27 & 34.19 & 42.47 & 22.87 & 43.59 & 53.35 \\
    E2VID-CLIP        &  5.95 & 14.72 & 19.91 & 12.11 & 23.03 & 27.95 \\
    CETA-Prompt-S & 14.67 & 33.51 & 42.44 & 23.16 & 44.15 & 54.51 \\
    CETA-BLIP2-S & 22.80 & 47.03 & 57.31 & \underline{29.03} & 52.63 & \underline{62.55} \\

    \midrule
    CEIA-S & \underline{24.75} & \underline{49.27} & \underline{58.55} & 28.95 & \underline{52.75} & 62.27 \\
    CEIA-L & \textbf{29.44} & \textbf{57.44} & \textbf{66.51} & \textbf{34.23} & \textbf{59.79} & \textbf{69.47} \\

    \bottomrule
  \end{tabular}
  }
\end{table}

\noindent \textbf{Results.} 
\cref{tab:e_i_retrieval} shows that our CEIA-L consistently outperforms EventCLIP and E2VID-CLIP under all metrics across both datasets. 
Specifically, on N-ImageNet, CEIA-L surpasses EventCLIP and E2VID-CLIP by 37.61\% and 38.73\% in terms of R@1 for event queries, respectively. 
% While these improvements may seem surprising, they are indeed reasonable. 
The underlying reason is that the contrastive loss we used is essential for multi-modal retrieval as it directly learns cross-modal similarity and alleviates the domain disparity of event and image data. 
When compared to CETA-Prompt and CETA-BLIP2, CEIA also holds overwhelming advantages because it directly aligns event-image data. For instance, CEIA-S outperforms CETA-Prompt-S and CETA-BLIP2-S by 32.81\% and 30.42\% in terms of R@1 for event queries on N-ImageNet. 
Additionally, we visualize the results of the image query on N-ImageNet in \cref{fig:visualization}, where the retrieved events have a very high degree of similarity to the input image query.

\subsection{Event-Text Retrieval}

\noindent \textbf{Metrics.} 
Similar to the event-image retrieval task, we reuse the recall at K (R@K) \cite{lee2018stacked} to evaluate the performance of the event-text retrieval task.

\noindent \textbf{Results.} As shown in \cref{tab:e_t_retrieval}, we report the results on our built event-text dataset N-ImageNet-BLIP2.
As can be seen, our CEIA-L outperforms EventCLIP and E2VID-CLIP by a large margin in both event query and text query. For example, CEIA-L achieves an 11.36\% improvement in terms of R@1 for text query compared to EventCLIP. Although CETA-BLIP2-S achieves slightly better results than our CEIA-S for text query, it's an unfair comparison as CETA-BLIP2-S employs the captions generated by BLIP2 for training, which have the same distribution as N-ImageNet-BLIP2.
% Overall, the results in \cref{tab:e_t_retrieval} demonstrate the effectiveness of our strategy, which leverages abundant paired event-image data as an alternative to overcome the lack of paired event-text data. 
In addition, we qualitatively show the results of CEIA in \cref{fig:visualization}. Even when the caption describes the relationship of multiple objects (the 4th row), CEIA is able to accurately retrieve the most correlated events.

% We follow the setting similar to event-image retrieval and conduct event-text retrieval experiments. In \cref{tab:e_t_retrieval}, we can see that ELIP also achieves significant improvements in both event query and text query on N-ImageNet-BLIP2. This highlights the success of our strategy, which aligns event and language modalities using the visual modality as a bridge.

\subsection{Domain Adaptation}

\begin{table}[tb]
  \caption{Quantitative results of domain adaptation.}
   \label{tab:domain_adaptation}
   \centering
  \setlength{\tabcolsep}{10pt}
  \renewcommand{\arraystretch}{1}
  \resizebox{0.8\linewidth}{!}{
   \begin{tabular}{c|cc|cc}
    \toprule

    \multirow{2}{*}{\centering Method}
    & \multicolumn{2}{c|}{N-ImageNet\cite{kim2021n}} & \multicolumn{2}{c}{N-Caltech101\cite{orchard2015converting}}\\
    \cmidrule(lr){2-3} \cmidrule(lr){4-5}

    & Acc1 & Acc5 & Acc1 & Acc5 \\

    \midrule
    EventCLIP \cite{wu2023eventclip}         & 21.92 & 43.24 & 69.22 & 80.46 \\
    E2VID-CLIP        & 10.86 & 23.88 & \underline{81.12} & \underline{93.76} \\
    CETA-Prompt-S & 21.32 & 38.58 & 70.83 & 83.10 \\
    CETA-BLIP2-S & 27.66 & 47.40 & 72.20 & 84.04 \\

    \midrule
    CEIA-S & \underline{43.64} & \underline{70.56} & 79.38 & 91.13 \\
    CEIA-L & \textbf{50.16} & \textbf{76.36} & \textbf{85.17} & \textbf{94.89} \\

    \bottomrule
  \end{tabular}
  }
\end{table}

\noindent \textbf{Setting.}
We conduct domain adaptation based on object recognition, which aims to validate the effectiveness of enhancing event-based understanding by transferring the knowledge of the frame-based vision. Specifically, We first train a classifier as the task network using labeled data from the image domain, and then directly transfer it to the event domain.

\noindent \textbf{Results.}
As observed in \cref{tab:domain_adaptation}, our CEIA-L consistently secures the top position on both N-ImageNet and N-Caltech101. Specifically, in terms of Acc1, CEIA-L outperforms E2VID-CLIP by 39.30\% on N-ImageNet and by 4.05\% on N-Caltech101. Moreover, compared to CETA-Prompt-S and CETA-BLIP2-S, our CEIA-S also exhibits its superiority, achieving significant increases for all metrics. These remarkable improvements demonstrate that CEIA effectively aligns event and image data within the same embedding space. Consequently, CEIA facilitates a smoother transfer of the task network trained in the image domain to the event domain, thereby enhancing the performance of domain adaptation.  
We believe that these performance gains can be transferred to other tasks and unlock the virtually unlimited image-based datasets for event-based vision, which will be our future work.

\begin{table}[tb]
  \caption{Ablation study on the effects of LoRA on event-text retrieval and domain adaptation.}
  \vspace{-2mm}
   \label{tab:wo_lora}
  \centering
  \setlength{\tabcolsep}{3pt}
  \renewcommand{\arraystretch}{1}
  \resizebox{1.0\linewidth}{!}{
  \begin{tabular}{c|ccc|ccc|cc|cc}
    \toprule

    \multirow{3}{*}{\centering Method}
    & \multicolumn{6}{c|}{Event-text retrieval} & \multicolumn{4}{c}{Domain adaptation}\\
    
    \cmidrule(lr){2-7} \cmidrule(lr){8-9} \cmidrule(lr){10-11} 
    
    & \multicolumn{3}{c|}{Event query} & \multicolumn{3}{c|}{Image query} & \multicolumn{2}{c|}{N-ImageNet\cite{kim2021n}} & \multicolumn{2}{c}{N-Caltech101\cite{orchard2015converting}} \\
    
    \cmidrule(lr){2-4} \cmidrule(lr){5-7} \cmidrule(lr){8-9} \cmidrule(lr){10-11}
    & R@1 & R@5 & R@10 & R@1 & R@5 & R@10 & Acc1 & Acc5 & Acc1 & Acc5\\

    \midrule
    w/o LoRA        & 28.99 & 55.75 & 64.95 & 31.23 & 57.95 & 67.79
                    & 48.24 & 74.02 & 83.82 & 94.46 \\
    w/ LoRA         & \textbf{29.44} & \textbf{57.44} & \textbf{66.51} & \textbf{34.23} & \textbf{59.79} & \textbf{69.47}
    & \textbf{50.16} & \textbf{76.36} & \textbf{85.17} & \textbf{94.89} \\

    \bottomrule
  \end{tabular}
  }
  \vspace{-4mm}
\end{table}

\section{Ablation Study}

\noindent \textbf{The Effectiveness of LoRA.} We compare two methods of training the event encoder: full finetuning and LoRA-based finetuning \cite{hu2021lora}. From \cref{tab:wo_lora}, we can observe that, LoRA-based finetuning consistently outperforms full finetuning across all metrics for event-text retrieval and domain adaptation tasks. These results demonstrate that LoRA can effectively preserve CLIP’s strong robustness and meanwhile avoid overfitting to the training datasets.

\noindent \textbf{LoRA Configuration.} In \cref{tab:LoRA}, we evaluate various LoRA \cite{hu2021lora} configurations as depicted in \cref{fig:method}. ``$r$'' represents the low intrinsic dimension of rank decomposition matrices. ``$\alpha$'' indicates the scaling degree applied to the outputs from the trainable weights, and ``Weight Type'' denotes which weight matrices in the event encoder are finetuned with LoRA. The experimental results demonstrate that adapting only $W_{q}$ and $W_{v}$ with a very small $r$ has already achieved competitive performance. Further increasing $r$ or adjusting LoRA with more weights does not lead to significant improvements. Additionally, we set $\alpha$ as twice $r$ to scale up the output from trainable weights, thereby further speeding up training.

\begin{figure}[tb]
  \centering
  \includegraphics[width=12.1cm]{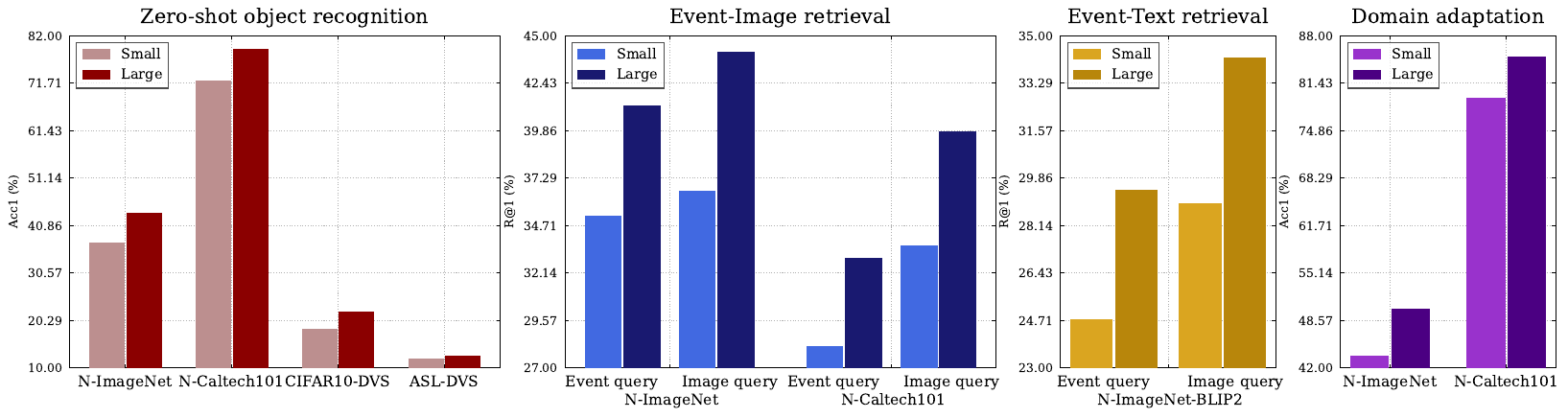}
  \caption{Comparison of training CEIA with different scale data.}
  \label{fig:scale}
\end{figure}

\noindent \textbf{Data Scalable Capability.}
As shown in \cref{fig:scale}, we can see that CEIA-L, which is trained on the larger-scale event-image pairs, achieves significantly better performance than CEIA-S across all benchmarks. This indicates that, leveraging more training data, CEIA can exhibit the flexibility to boost performance, ensuring its scalable capability. Therefore, larger-scale event-image pretraining is an exciting direction for future work.

% \begin{table}[tb]
%   \caption{Ablation study on event representations on N-ImageNet (S).}
%   \label{tab:representations}
%   \centering
%   \setlength{\tabcolsep}{8pt}
%   \renewcommand{\arraystretch}{1}
%   \resizebox{0.5\linewidth}{!}{
%   \begin{tabular}{c|cc}
%     \toprule
    
%     Representations &  Acc1 & Acc5  \\
%     \midrule        
%     DiST            & 33.34 & 57.64 \\
%     Time Surface    & 34.84 & 58.74 \\
%     Voxel           & 30.91 & 54.24 \\
%     Gray            & \underline{35.94} & \underline{60.31} \\
%     R\&B            & \textbf{37.25} & \textbf{61.60} \\
%     \bottomrule
%   \end{tabular}
%   }
% \end{table}

\noindent \textbf{Event Representations.} 
The results in \cref{tab:representations} show the ablation results of different event representations.
% demonstrate the optimal representation to choose when transforming events into frames.
Compared with the commonly-used DiST \cite{kim2021n}, Time Surface \cite{lagorce2016hots}, Voxel \cite{zhu2019unsupervised}, and Gray \cite{wu2023eventclip}, the red-blue color map (referred to as R-B) \cite{wu2023eventclip} leads to the best recognition accuracy.
% Specifically, we first adopted the red-blue color map (referred to as R-B) to introduce polarity information, thereby resulting in a 1.31\% improvement. Furthermore, we attempted to use Voxel \cite{zhu2019unsupervised} to introduce temporal information or DiST \cite{kim2021n} to suppress noisy events, but both approaches led to worse results than anticipated. 
We speculate that these worse results may be due to larger differences between these representations and natural images used by CLIP.

\begin{table}[t!]
\begin{floatrow}
% %%%%%%%%%%%%%%%%%%%%%%%%%%%%%%%%%
% tab1
% %%%%%%%%%%%%%%%%%%%%%%%%%%%%%%%%%
\setlength{\tabcolsep}{6pt}
\renewcommand{\arraystretch}{1}
\capbtabbox{
\resizebox{0.95\linewidth}{!}{
\begin{tabular}{c|ccccc}
    \toprule
            
    Rank $r$        &   16          & 32            & 64            & 16            & 16 \\
    LoRA $\alpha$   &   16          & 32            & 64            & 32            & 32 \\
    \makecell[c]{Weight\\Type}     & \makecell[c]{$W_{q},$\\$W_{v}$} & \makecell[c]{$W_{q},$\\$W_{v}$} & \makecell[c]{$W_{q},$\\$W_{v}$} & \makecell[c]{$W_{q},$\\$W_{v}$} & \makecell[c]{$W_{q},W_{k},$\\$W_{v},W_{o}$}\\
    \midrule
    Acc1            & 36.28         & 36.80          & 36.80          & \underline{37.25}         & \textbf{37.33} \\
    \bottomrule
  \end{tabular}
}
}{\caption{Ablation study on LoRA configuration on N-ImageNet-S.}
 \label{tab:LoRA}}
\vspace{-4mm}

% %%%%%%%%%%%%%%%%%%%%%%%%%%%%%%%%%
% tab2
% %%%%%%%%%%%%%%%%%%%%%%%%%%%%%%%%%
\setlength{\tabcolsep}{12pt}
\renewcommand{\arraystretch}{1}
\capbtabbox{
\resizebox{0.9\linewidth}{!}{
\begin{tabular}{c|cc}
    \toprule
    
    Representations &  Acc1 & Acc5  \\
    \midrule        
    DiST \cite{kim2021n}           & 33.34 & 57.64 \\
    Time Surface \cite{lagorce2016hots}     & 34.84 & 58.74 \\
    Voxel \cite{zhu2019unsupervised}          & 30.91 & 54.24 \\
    Gray \cite{wu2023eventclip}  & \underline{35.94} & \underline{60.31} \\
    R-B \cite{wu2023eventclip}   & \textbf{37.25} & \textbf{61.60} \\
    \bottomrule
  \end{tabular}
}
}{\caption{Ablation study on event representations on N-ImageNet-S.}
 \label{tab:representations}}

\end{floatrow}
\end{table}

\section{Conclusion}
In this paper, we propose CEIA, an effective framework to adapt CLIP to event data. We provide a novel perspective of focusing on learning to align event and image data as an alternative, thus overcoming the challenge posed by the shortage of event-text datasets. We thoroughly evaluate CEIA on four applications: object recognition, event-image retrieval, event-text retrieval, and domain adaptation. The state-of-the-art results show that CEIA not only enhances open-world understanding but also opens the door to more event-based multi-modal understanding tasks. Furthermore, CEIA's significant scalability under abundant event-image pairs also opens up the possibility to introduce the first event-based Large Vision Model, which will be our future work.

\bibliographystyle{splncs04}
\bibliography{main}
\end{document}